\documentclass[10pt,twocolumn,letterpaper]{article}

\usepackage{wacv}
\usepackage{normalpkgs} 

\def\wacvPaperID{962} 

\wacvfinalcopy 

\ifwacvfinal
\fi

\ifwacvfinal
\usepackage[pagebackref=true,breaklinks=true,colorlinks,bookmarks=false]{hyperref}
\else
\usepackage[pagebackref=true,breaklinks=true,colorlinks,bookmarks=false]{hyperref}
\fi

\ifwacvfinal
\fi

\begin{document}

\title{Cross-Domain Latent Modulation for Variational Transfer Learning}

\author{Jinyong Hou, Jeremiah D. Deng, Stephen Cranefield, Xuejie Ding  \\
Department of Information Science, University of Otago\\
{\tt\small robert.hou@postgrad.otago.ac.nz, \{jeremiah.deng,stephen.cranefield,emily.ding\}@otago.ac.nz}
\and
}

\maketitle
\thispagestyle{empty}

\begin{abstract}
    We propose a cross-domain latent modulation mechanism within a variational autoencoders (VAE) framework to enable improved transfer learning. Our key idea is to procure deep representations from one data domain and use it as perturbation to the reparameterization of the latent variable in another domain. Specifically, deep representations of the source and target domains are first extracted by a unified inference model and aligned by employing gradient reversal. Second, the learned deep representations are cross-modulated to the latent encoding of the alternate domain. The consistency between the reconstruction from the modulated latent encoding and the generation using deep representation samples is then enforced in order to produce inter-class alignment in the latent space. We apply the proposed model to a number of transfer learning tasks including unsupervised domain adaptation and image-to-image translation. Experimental results show that our model gives competitive performance. 
\end{abstract}


\section{Introduction} \label{sec:intro}

In machine learning, one can rarely directly apply a pre-trained model to a new dataset or a new task, as the performance of a learned model often plunges significantly for the new data which may have significant sampling bias or even belong to different distributions. Transfer learning can help us utilize the learned knowledge from a previous domain (the `source') to improve performance on a related domain or task (the `target')~\cite{Pan2010,Tan2018,Weiss2016}. 

From the perspective of probabilistic modeling ~\cite{Liu2017,Wang2017a,Schonfeld2019}, the key challenge in achieving cross-domain transfer is to learn a joint distribution of data from different domains. Once the joint distribution is learned, it can be used to generate the marginal distribution of the individual domains~\cite{Kingma2013a,Liu2017}. Under the variational inference scenario, an inferred joint distribution is often applied to the latent space. Due to the coupling theory, inferring the joint distribution from the marginal distributions of different domains is a highly ill-posed problem~\cite{Lindvall2002Prob}. To address this problem, UNIT~\cite{Liu2017} makes an assumption that there is a shared latent space for the two domains. Usually, this can be achieved by applying the adversarial strategy to the domains' latent spaces. Another line of research focuses on the use of a complex prior to improve the representation performance for the input data~\cite{Mahajan2020,Tomczak2018,Hoffman2016}. However, the previous works neglect the role of the generation process for the latent space which could be helpful for cross-domain transfer scenarios.  

In this paper, we propose a novel latent space reparameterization method, and employ a generative process to cater for the cross-domain transferability. Specifically, we incorporate a cross-domain component into the reparameterization transformation, which builds the connection between the variational representations and domain features in a cross-domain manner. The generated transfer latent space is further tuned by domain-level adversarial alignment and domain consistency between images obtained through reconstruction and generations. We apply our model to the homogeneous transfer scenarios, such as unsupervised domain adaptation and image-to-image translation. The experimental results show the efficiency of our model. 

The rest of the paper is organized as follows. In Section~\ref{sec:related_work}, some related work is briefly reviewed. In Section~\ref{sec:model}, we outline the overall structure of our proposed model and develop the learning metrics with defined losses. The experiments are presented and discussed in Section~\ref{sec:experiments}. We conclude our work in Section~\ref{sec:conclusion}, indicating our plan of future work.

\section{Related Work} \label{sec:related_work}
\noindent
\textbf{Latent space manipulation:} As discussed above, for a joint distribution, manipulation of the latent space is common~\cite{Larsen2016,Liu2017,Liu2018b} for the cross-domain adaptation situations. One approach focuses on a shared latent space, where the latent encodings are regarded as common representations for inputs across domains. Some adversarial strategy is usually used to pool them together so that the representations are less domain-dependent. For the variational approach, works in~\cite{Mahajan2020,Ziegler2019,Kingma2016} adopt complex priors for multi-modal latent representations, while other works~\cite{Liu2017,Schonfeld2019,Liu2018b} still assume a standard Gaussian prior. Another aproach is to use disentangled latent representations where the latent encoding is divided into some defined parts (e.g.~style and content parts), then the model learns separated representations and swaps them for the transfer~\cite{Gonzalez-garcia2018,Zhang2019d,Feng2018a,Lee2018}. Our method is different from these approaches. In our model, learned auxiliary deep representation is used to generate perturbations to the latent space through a modified reparameterization using variational information from the counterpart domain. It helps generate cross-domain image translation. 
The transfer is carried out by a reparameterization transformation, using statistical moments retaining specific information for one domain, and deep representation providing information from another domain. 

\textbf{Varied Homogeneous transfer tasks:} The manipulation on the latent space is often interwoven with the homogeneous image transfer together, such as unsupervised domain adaptation and image translation~\cite{Naseer2019,Noguchi2019,Choi2018}. In the domain separation networks~\cite{Bousmalis2016}, separate encoding modules are employed to extract the invariant representation and domain-specific representations from the domains respectively, with the domain-invariant representations being used for the domain adaptation.  References~\cite{Bousmalis2017,Sankaranarayanan2018,Hoffman2018} transfer the target images into source-like images for domain adaptation.  References~\cite{Liu2017,Zhu2017,Kim2017} map the inputs of different domains to a single shared latent space, but cycle consistency is required for the completeness of the latent space. The UFDN~\cite{Liu2018b} utilizes a unified encoder to extract the multi-domain images to a shared latent space, and the latent domain-invariant coding is manipulated for image translation between different domains. 

In contrast, we adopt the pixel-level adaptation between domains from the cross-domain generation, but the proposed model can also be used at the feature-level due to the latent space alignment. Our model also has a unified inference model, but the consistency is imposed in a straightforward way, with reduced computational complexity. 



\section{Proposed Model} \label{sec:model}

\subsection{Problem setting}
\label{subsec:problemsetting}
Let $\mathcal{X} \subset \mathbb{R}^d$ be a $d$-dimensional data space, and $\bm X = \{\bm x_1, \bm x_2, \ldots, \bm x_n\} \in \mathcal{X}$ the sample set with marginal distribution $p(\bm X)$. The source domain is denoted by a tuple $(\mathcal{X}_s, p(\bm X_s))$, and the target domain by $(\mathcal{X}_t, p(\bm X_t))$. In our paper, we consider the homogeneous transfer with domain shift, i.e. $\mathcal{X}_s \approx \mathcal{X}_t$, but $p(\bm X_s) \ne p(\bm X_t)$. For the unsupervised pixel-level domain adaptation scenario, the label set is $\bm Y=\{y_1, y_2, \ldots y_n\} \in \mathcal{Y}$ ($\mathcal{Y}$ is the label space), and a task $\mathcal{T} = p(\bm Y|\bm X)$ is considered too. However, only the source domain's label set $\bm Y_s$ is available during transfer learning.

\subsection{Transfer Latent Space}
\label{subsec:transfer_latent_space}

As any given marginal distribution can be yielded by an infinite number of joint distributions, we need to build an inference framework with some constraints. Under the variational autoencoder (VAE) framework, the latent space is one of the manipulation targets. We propose the transfer latent space as follows.

\theoremstyle{definition}
\begin{definition}{\textbf{Transfer Latent Space $\ddot{\bm Z}$.}}
  Let $\bm x_s \in \bm {X}_s$, $\bm x_t \in \bm {X}_t$ be the domain samples. Let us have a map $f$ that extracts domain information $\bm\Omega$ and a feature representation $\bm h$ given an input $\bm x$:
  \[ f: \bm x \longrightarrow (\bm\Omega, \bm h), \bm x \in \bm X_s \cup \bm X_t.\] 
  Suppose we construct a transfer map $\mathcal{G}$ that generates a latent variable $\bm{\ddot{z}}$ from $\bm\Omega$ and $\bm h$ with domain crossovers:
  \[ \bm{\ddot{z}}_{st}=\mathcal{G}(\bm\Omega_s,\bm h_t), \]
  \[ \bm{\ddot{z}}_{ts}=\mathcal{G}(\bm\Omega_t,\bm h_s). \]
  The joint space formed by $\bm{\ddot{z}}_{st}$ and $\bm{\ddot{z}}_{ts}$ samples is defined as a \textit{transfer latent space}, denoted by $\ddot{\bm Z}$. 
\end{definition}

The transfer latent space is intended to become a ``mixer'' for the two domains, as the resulted latent variables are under cross-domain influences. Hence the transfer latent space can be regarded as a generalization of the latent space.

\subsection{Framework}
\label{subsec:freamework}

\begin{figure*}[tbhp]
    \centering
    \includegraphics[width=0.75\textwidth]{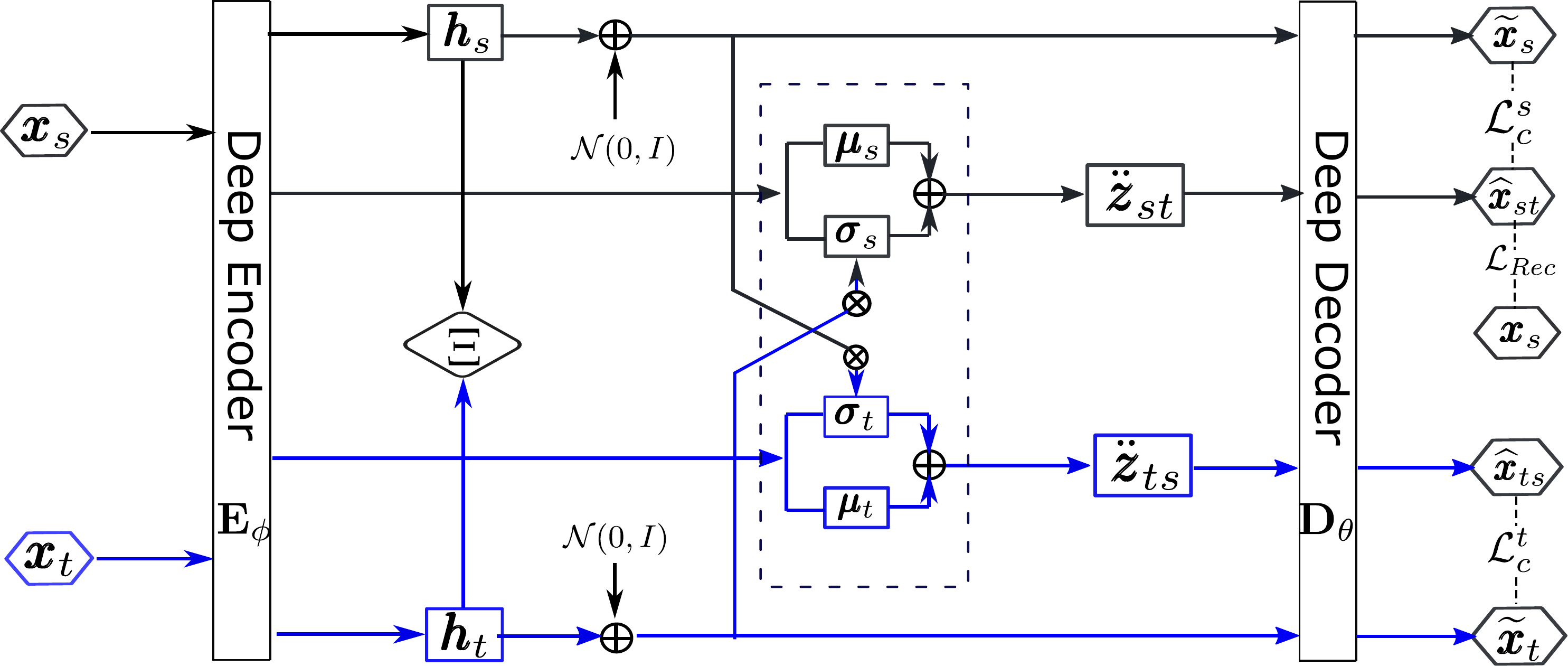}
    \caption{\small Architectural view of the proposed model. It encourages an image from target domain (blue hexagon) to be transformed to a corresponding image in the source domain (black hexagon). The transfer latent distributions $p(\bm {\ddot{z}}_{ts}|\bm x_t, \bm h_s)$ and $p(\bm{\ddot{z}}_{st}|\bm x_s, \bm h_t)$ are learned which are used to generate corresponding images by the desired decoder. The deep representations are integrated into the reparameterization transformation with standard Gaussian auxiliary noise. Blue lines are for the target domain and black ones are for the source domain.}
    \label{fig:model_sche}
\end{figure*}

Our framework is shown in Fig.~\ref{fig:model_sche}. In our framework, we build the cross-domain generation by a unified inference model $\mathbf{E}_{\phi}(\cdot)$ (as an implementation of the map $f$) and a generative model for the desired domain $\mathbf{D}_{\theta}(\cdot)$, e.g., the source domain in our model. 
A discriminator $\Xi$ is utilized for the adversarial training. We use the terms ``inference model'' and ``encoder'' for $\mathbf{E}_{\phi}(\cdot)$, and ``generative model'' and ``decoder'' for  $\mathbf{D}_{\theta}(\cdot)$ interchangeably. 

As discussed in section~\ref{subsec:transfer_latent_space}, under the variational framework, the domain information $\bm \Omega$ (here we remove the domain subscript for simplicity) is usually the pair $(\bm \mu, \bm \sigma)$. Let $\bm h'$ be the flattened activations of the last convolution layer in $\mathbf{E}_u$. Then, following the treatment in~\cite{Kingma2014}, $\bm \mu$ and $\bm \sigma$ can be obtained by $\bm \mu = \mathbf{W}_{\bm \mu}\bm h' + \bm b_{\bm \mu}$ and $\bm \sigma = \mathbf{W}_{\bm \sigma}\bm h' + \bm b_{\bm \sigma}$, where $\mathbf{W}_{\bm \mu}, \mathbf{W}_{\bm \sigma}, \bm b_{\bm \mu}, \bm b_{\bm \sigma}$ are the weights and biases for $\bm \mu$ and $\bm \sigma$.

From our observations, both shallow (e.g.~PCA features) and deep representations can be used to obtain domain information $\bm h$, in our end-to-end model we use the latter. We choose the high-level activation of the last convolutional layer, i.e., $\bm h = \text{sigmoid}(\mathbf{W}_{\bm h}\bm h' + \bm b_{\bm h})$ as the deep representation~\cite{Yosinski2014}, where $\mathbf{W}_{\bm h}, \bm b_{\bm h}$ are the weights and biases for the deep abstractions. 

Having obtained the domain information $\bm\Omega$ and deep representation $\bm h$, a natural choice for the transfer map $\mathcal{G}$ is through reparameterization. Here we propose a modified reparameterization trick to give the sampling from 
the transfer latent space as follows:
\begin{equation}
\bm{\ddot{z}}_{st} = \mathcal{G}((\bm\mu_s, \bm\sigma_s), \bm h_t) = \bm\mu_s + \bm\sigma_s \odot (\gamma_1\bm h_t + \gamma_2\bm\epsilon), 
\label{eq:zs_addition_model}
\end{equation}
and
\begin{equation}
\bm{\ddot{z}}_{ts} = \mathcal{G}((\bm\mu_t, \bm\sigma_t), \bm h_s) = \bm\mu_t + \bm\sigma_t \odot (\gamma_1\bm h_s + \gamma_2\bm\epsilon),
\label{eq:zt_addition_model}  
\end{equation}
where $\bm h_s$ ($\bm h_t$) is the sample of the deep representation space $\mathcal{H}_s$ ($\mathcal{H}_t$); $\bm \mu_s$ and $\bm\sigma_s$ ($\bm \mu_t$ and $\bm\sigma_t$) are the mean and standard deviation of the approximate posterior for the source (target) domain; $\gamma_1, \gamma_2>0$ are trade-off hyperparameter to balance the deep feature modulation and the standard Gaussian noise $\bm\epsilon$; and $\odot$ stands for the element-wise product of vectors. Therefore, the auxiliary noise in VAE resampling is now a weighted sum of a deep representation from the other domain and Gaussian noise, different from the standard VAE framework. Because the modified reparameterization allows a domain's deep representation to get modulated into another domain's latent representation, we call our model ``Cross-Domain Latent Modulation'', or CDLM for short.

Now we have obtained the transfer encodings by a unified encoder. Following the probabilistic encoder analysis~\cite{Kingma2014}, a shared inference model confines the latent variables into the same latent space. But this cannot guarantee them to be aligned. To pull the domains close, an adversarial strategy~\cite{Ganin2016, Tzeng2017} should be used for the alignment. The gradient reversal layer~\cite{Ganin2016} is used in our model, by which adversarial learning is introduced to learn transferable features that are robust to domain shift. The adversarial alignment between  $\mathcal{H}_s$ and $\mathcal{H}_t$ is for domain-level. 

Furthermore, for better interpretation of the modulated reparameterization, let $\bm h_t \sim p(\mathcal{H}_t) = \mathcal{N}(\bm \mu_{\bm h_t}, \bm \sigma_{\bm h_t})$. Then for $\bm{\ddot{z}}_{st}$, we have $\bm{\ddot{z}}_{st} \sim \mathcal{N}(\bm{\ddot{z}}_{st}; \bm \mu_{st}, \bm \sigma_{st}^2\mathbf{I})$. For the $i$-th element of the distribution moments are given as follows:
\begin{equation}  \label{eq:new_mu}
\begin{split}
    \mu_{st}^i &= \mathbb{E}\{\mu^i_s + \sigma_s^i(\gamma_1 h_t^i + \gamma_2 \epsilon^i)\}\\
    &= \mu_s^i + \gamma_1 \sigma_s^i \mu_{h_t}^i.
\end{split}
\end{equation}
\begin{equation}  \label{eq:new_sigma}
    \begin{split}
        \mathbf{Var}(\ddot{z}_{st}^i)&=\mathbb{E}\{(\sigma_s^i(\gamma_1 h_t^i + \gamma_2\epsilon^i) - \gamma_1\sigma_s^i\mu_{h_t}^i)^2\} \\
        &= (\sigma_s^i)^2 [\gamma_1^2(\sigma_{h_t}^i)^2 + \gamma_2^2].
    \end{split}
\end{equation}

Therefore, the $\mu_{st}^i$ and $\sigma_{st}^i$ are
\begin{equation}  \label{eq:new_mu_sigma}
\begin{gathered}
    \mu_{st}^i = \mu_s^i + \gamma_1 \sigma_s^i \mu_{\bm h_t}^i \\
    \sigma_{st}^i = \sigma_s^i \sqrt{\gamma_1^2(\sigma_{\bm h_t}^i)^2 + \gamma_2^2}.
\end{gathered}
\end{equation}

Here, it is reasonable to assume $\bm \mu_{\bm h_t} \approx \bm \mu_t$ and $\bm \sigma_{\bm h_t} \approx \bm \sigma_t$ when the training is finished. With a practical setting of $\gamma_1\gg\gamma_2$, and in effect $\sigma_s^i=1$, Eq.~(\ref{eq:new_mu_sigma}) can be further simplified to $\mu_{st}^i=\mu_s^i+\gamma_1\mu_t^i$,
and $\sigma_{st}^i=\gamma_1\sigma_s^i\sigma_t^i$. Then we can see that $\bm \mu_{st}$ can be regarded as a location shift of $\bm\mu_s$ under the influence of $\bm \mu_t$, which helps reduce the domain gap; $\bm\sigma_{st}$ can be taken as a recoloring of $\bm\sigma_s$ under the influence from the target. The formulation of $\bm{\ddot{z}}_{ts}$ can be similarly interpreted. These modulated encodings are hence constructed in a cross-domain manner. 

Next, we apply the consistency constraint to the transfer latent space with modulation  for further inter-class alignment. It has been found that  consistency constraints preserve class-discriminative information~\cite{Schonfeld2019,Kim2017,Ghifary2016}. For our model, the consistency is applied to the reconstructions from modulated encodings and the corresponding generations from deep representations. Let $\mathbf{D}_{\theta}(\cdot)$ be the generative model for domain image generation from the transfer latent space. The consistency requirements are
\begin{equation}
  \begin{gathered}
    \mathbf{D}_{\theta}(\bm{\ddot{z}}_{st}) = \mathbf{D}_{\theta}(\gamma_1\bm h_s + \gamma_2 \bm \epsilon) \\
    \mathbf{D}_{\theta}(\bm{\ddot{z}}_{ts}) = \mathbf{D}_{\theta}(\gamma_1 \bm h_t + \gamma_2 \bm \epsilon),
  \end{gathered}
\end{equation}
where $\mathbf{D}_{\theta}(\bm{\ddot{z}}_{st})$ is the reconstruction of the source ($\widehat{\bm x}_{st}$), $\mathbf{D}_{\theta}(\bm{\ddot{z}}_{ts})$ is for the target ($\widehat{\bm x}_{ts}$). $\mathbf{D}_{\theta}(\cdot)$ can also function as a generative model, generating $\widetilde{\bm x}_s = \mathbf{D}_{\theta}(\gamma_1\bm h_s + \gamma_2 \bm \epsilon)$ and $\widetilde{\bm x}_{t} = \mathbf{D}_{\theta}(\gamma_1 \bm h_t + \gamma_2 \bm \epsilon)$ for the source and target domain respectively. Also, the consistencies can guide the encoder to learn the representations from both domains. 

Finally, a desired marginalized decoder, e.g. the source decoder, is trained to map the target images to be source-like. We render the target's structured generation $\widetilde{\bm x}_{ts}$ for the test mode. For this end, we do not need the source to be taken into account for the test. That means a test image from the target domain $\bm x_t^i$ first passes through the inference model and obtains its deep feature $\bm h_t^i$. Then it is fed into the generation model to generate an image with source style but keep its own class. That is to make the marginal distribution $p(\widetilde{\bm x}_{ts}^i) \approx p(\bm x_s^j)$, but keep its class $y_t^i$.

\subsection {Learning}
\label{subsec:learning}

Our goal is to update the variational parameters to learn a joint distribution and the generation parameters for the desired marginal distribution. Since the latent variables are generated with inputs from both domains, we have a modified formulation adapted from the plain VAE:
\begin{equation}
\resizebox{0.49\textwidth}{!}{$
\begin{aligned}
  \log p(\bm{x}_s, \bm{x}_t) - \mathrm{KL}(q_\phi(\bm{\ddot{z}}|\bm{x}_s, \bm{x}_t)\|p(\bm{\ddot{z}}|\bm{x}_s, \bm{x}_t))
  &= \mathbb{E}[\log p(\bm{x}_s, \bm{x}_t|\bm{\ddot{z}})] \\
  &-\mathrm{KL}(q_\phi(\bm{\ddot{z}}|\bm{x}_s, \bm{x}_t)\|p(\bm{\ddot{z}})),
\end{aligned}$}
\end{equation}
where $\mathrm{KL}(\cdot)$ is the Kullback-Leibler divergence, and the transfer latent variable $\bm{\ddot{z}}$ can be either $\bm{\ddot{z}}_{st}$ or $\bm{\ddot{z}}_{ts}$. Minimizing $\mathrm{KL}(q_\phi(\bm{\ddot{z}}|\bm{x}_s, \bm{x}_t)\|p(\bm{\ddot{z}}|\bm{x}_s, \bm{x}_t))$ is equivalent to maximizing the variational evidence lower bound (ELBO) $\mathcal{L}(\bm x_s, \bm x_t, \bm\theta, \bm\phi)$:
\begin{equation} \label{eq:ELBO_kl_model}
\resizebox{0.49\textwidth}{!}{$
  \mathcal{L}(\bm x_s, \bm x_t, \bm\theta, \bm\phi)  = \mathbb{E}_{q_\phi}[\log p_\theta(\bm{x}_s, \bm{x}_t|\bm{\ddot{z}})]
  -\mathrm{KL}(q_\phi(\bm{\ddot{z}}|\bm{x}_s, \bm{x}_t)\|p(\bm{\ddot{z}})),$}
\end{equation}
where the first term corresponds to the reconstruction cost ($\mathcal{L}_{\mathit{Rec}}$), 
and the second term is the K-L divergence between the learned latent probability and the prior (specified as $\mathcal{N}(0, \bm{I})$) ($\mathcal{L}_{\mathit{KL}}$). Considering the reconstruction of $\bm{x}_s$, and the K-L divergence for both $\bm{\ddot{z}}_{st}$ and $\bm{\ddot{z}}_{ts}$, we have
\begin{equation}
\begin{split}
\mathcal{L}(\bm x_s, \bm x_t, \bm\theta, \bm\phi) &= \mathbb{E}_{\bm{\ddot{z}} \sim q(\bm{\ddot{z}}_{st}|\bm x_s, \bm x_t)}[\log p_{\bm \theta}(\bm x_s|\bm{\ddot{z}}_{st})] \\
&- \mathrm{KL}(\log q_{\bm\phi}({\bm{\ddot{z}}_{st}|\bm x_s, \bm x_t})\|p(\bm z)) \\
& - \mathrm{KL}(\log q_{\bm\phi}({\bm{\ddot{z}}_{ts}|\bm x_t, \bm x_s})\|p(\bm z)). \\
\end{split}
\label{eq:vae}
\end{equation}

To align the deep representations of the source and target domains, an adversarial strategy is employed to regularize the model. The loss function is given by
\begin{equation}
\resizebox{0.49\textwidth}{!}{$
  \mathcal{L}_{adv}= \mathbb{E}_{\bm h_s \sim p(\bm h_s|\bm x_s)}[\log \Xi(\bm h_s)] + \mathbb{E}_{\bm h_t \sim p(\bm h_t|\bm x_t)}[\log (1 - \Xi(\bm h_t))],$}
\label{eq:adv}
\end{equation}
where $\Xi(\cdot)$ is the discriminator to predict from which domain the deep representation feature is. 


From the analysis in Section~\ref{subsec:transfer_latent_space}, we can introduce a pairwise consistency between the reconstruction and the generation for the source and the target in an unsupervised manner respectively. The consistencies regularization improve the inter-class alignment. For the consistency loss $\mathcal{L}_{c}$, both the $l^1$ and $l^2$-norm penalty can be used to regularize the decoder. Here we simply use MSE. Let $\mathcal{L}_c^s$ and $\mathcal{L}_c^t$ be the consistency for the domains respectively. $\mathcal{L}_c$ is given as a combination of these two components, weighted by two coefficients $\beta_1$ and $\beta_2$, respectively:
\begin{equation}
\resizebox{0.49\textwidth}{!}{$
\begin{aligned}
\mathcal{L}_{c} &= \beta_1 \mathcal{L}_c^s + \beta_2 \mathcal{L}_c^t \\
&= \beta_1 (\mathbb{E}_{\bm{\ddot{z}} \sim q(\bm{\ddot{z}}|\bm x_s, \bm x_t)}[\log p(\widehat{\bm x}_s|\bm{\ddot{z}}_{st})] - \mathbb{E}_{\bm h_s \sim p(\bm h_s|\bm x_s)}[\log p(\widetilde{\bm x}_s|\bm h_s)])^2 \\
&+ \beta_2 (\mathbb{E}_{\bm{\ddot{z}} \sim q(\bm{\ddot{z}}|\bm x_s, \bm x_t)}[\log p(\widehat{\bm x}_{ts}|\bm{\ddot{z}}_{ts})] - \mathbb{E}_{\bm h_t \sim p(\bm h_t|\bm x_t)}[\log p(\widetilde{\bm x}_{ts}|\bm h_t)])^2. 
\end{aligned}$}
\label{eq:loss_consistency}
\end{equation}   

Then, the variational parameters $\bm \phi$ and generation parameters $\bm \theta$ are updated by the following rules:
\begin{equation}
  \begin{gathered}
    \bm \phi \leftarrow \bm \phi - \eta_1 \nabla ( \mathcal{L}_{\mathit{adv}} + \lambda_1 \mathcal{L}_{\mathit{KL}} + \lambda_2 \mathcal{L}_{\mathit{Rec}}) \\
    \bm \theta \leftarrow \bm \theta - \eta_2 \nabla (\mathcal{L}_{\mathit{Rec}} + \mathcal{L}_{c}),
  \end{gathered}
  \label{eq:update}
\end{equation}
where $\eta_1, \eta_2$ are the learning rates. Note, that only data from the desired domain (the source) are used to train the reconstruction loss. The $\mathrm{KL}$ items approximate the transfer latent space to their prior. Hyperparameters $\lambda_1$, $\lambda_2$ are used to balance the discriminator loss and reconstruction loss.


\section{Experiments} \label{sec:experiments}

\begin{table*}[]
    \centering
    \caption{\small Mean classification accuracy comparison. The ``source only" row is the accuracy for target without domain adaptation training only on the source. The ``target only" is the accuracy of the full adaptation training on the target. For each source-target task the best performance is in bold}
    \resizebox{0.8\textwidth}{!}{
        \begin{tabular}{|l|c|c|c|c|c|c|c|c|c|}
            \hline
            Source & MNIST & USPS & MNIST& MNISTM & Fashion & Fashion-M & Linemod 3D\\
            Target & USPS  & MNIST &MNISTM & MNIST& Fashion-M & Fashion & Linemod Real\\
            \hline\hline
            Source Only & 0.634 & 0.625 & 0.561  & 0.633  & 0.527 & 0.612 & 0.632 \\
            \hline\hline
            DANN~\cite{Ganin2016} & 0.774 & 0.833 & 0.766  &  0.851   &  0.765  & 0.822 & 0.832  \\
            CyCADA~\cite{Hoffman2018} & 0.956  & 0.965 &  0.921  &  0.943  & 0.874 & 0.915 &0.960 \\
            GtA~\cite{Sankaranarayanan2018} & 0.953  & 0.908  &  0.917 &  0.932 & 0.855 & 0.893 & 0.930  \\
            CDAN~\cite{Long2018} & 0.956 & 0.980 & 0.862 & 0.902 & 0.875 & 0.891 & 0.936  \\
            PixelDA~\cite{Bousmalis2017} & 0.959 & 0.942 &  0.982 &  0.922  & 0.805   &  0.762 & \textbf{0.998} \\
            UNIT~\cite{Liu2017} & 0.960 & 0.951  & 0.920  & 0.932 & 0.796   &  0.805 & 0.964 \\
            \hline\hline
            CDLM ($\widetilde{\bm x}_{ts}$) & \textbf{0.961}  & \textbf{0.983} & \textbf{0.987}   & \textbf{0.962}  &   \textbf{0.913} & \textbf{0.922} & 0.984  \\
            \hline \hline
            Target Only & 0.980 &  0.985 & 0.983  & 0.985  &  0.920  &  0.942 & 0.998   \\
            \hline  
    \end{tabular}}
    \label{table:accuracy}
\end{table*}

We conducted extensive evaluations of CDLM in two homogeneous transfer scenarios including unsupervised domain adaptation and image-to-image translation. During the experiments, our model was implemented using TensorFlow~\cite{Abadi2016}. The structures of the encoder and the decoder adopt those of UNIT~\cite{Liu2017} which perform well for image translation tasks. A two-layer fully connected MLP was used for the discriminator. SGD with momentum was used for updating the variational parameters, and Adam for updating generation parameters. The batch size was set to 64. During the experiments, we set  $\gamma_1=1.0,\gamma_2=0.1$,   $\lambda_1=\lambda_2=0.0001$, $\beta_1=0.1$ and $\beta_2=0.01$. For the datasets, we considered a few popular benchmarks, including MNIST~\cite{LeCun1998}, MNSITM~\cite{Ganin2016}, USPS~\cite{LeCun1989}, Fashion-MNIST~\cite{Xiao2017}, Linemod~\cite{Hinterstoisser2012,Wohlhart2015}, Zap50K-shoes~\cite{Yu2017} and CelebA~\cite{Liu2015,Liu2018}.

\subsection {Datasets}
\label{supp:datasets}
We have evaluated our model on a variety of benchmark datasets. They are described as follows. 


\noindent
\textbf{MNIST:} MNIST handwritten dataset~\cite{LeCun1998} is a very popular machine learning dataset. It has a training set of 60,000 binary images, and a test set of 10,000. There are 10 classes in the dataset. In our experiments, we use the standard split of the dataset. MNISTM~\cite{Ganin2016} is a modified version for the MNIST, with random RGB background cropped from the Berkeley Segmentation Dataset\footnote{URL https://www2.eecs.berkeley.edu/Research/Projects/CS/vision/bsds/}.

\noindent
\textbf{USPS:} USPS is a handwritten zip digits datasets~\cite{LeCun1989}.  It contains 9298 binary images ($16\times16$), 7291 of which are used as the training set, while the remaining 2007 are used as the test set. The USPS samples are resized to $28\times28$, the same as MNIST.  

\noindent
\textbf{Fashion:} Fashion~\cite{Xiao2017} contains 60,000 images for training, and 10,000 for testing. All the images are grayscale, $28\times28$ in size space. In addition, following the protocol in~\cite{Ganin2016}, we add random noise to the Fashion images to generate the FashionM dataset, with random RGB background cropped from the Berkeley Segmentation Dataset. 

\noindent
\textbf{Linemod 3D images} 
Following the protocol of~\cite{Bousmalis2017}, we render the LineMod~\cite{Hinterstoisser2012,Wohlhart2015} for the adaptation between synthetic 3D images (source) and real images (target). The objects with different poses are located at the center of the images. The synthetic 3D images render a black background and a variety of complex indoor environments for real images. We use the RGB images only, not the depth images. 
 
\noindent
\textbf{CelebA:} CelebA~\cite{Liu2015} is a large celebrities face image dataset. It contains more than 200k images annotated with 40 facial attributes. We select 50K images randomly, then transform them to sketch images followed the protocol of~\cite{Liu2018}. The original and sketched images are used for translation. 

\noindent
\textbf{UT-Zap50K-shoes:} This dataset~\cite{Yu2017} contains 50K shoes images with 4 different classes. During the translation, we get the edges produced by canny detector.

\subsection{Unsupervised Domain Adaptation}
\label{subsec:UDA}

We applied our model to unsupervised domain adaptation, adapting a classifier trained using labelled samples in the source domain to classify samples in the target domain. For this scenario, only the labels of the source images were available during training. 
We chose DANN~\cite{Ganin2016} as the baseline, but also compared our model with the state-of-the-art domain adaptation methods: Conditional Domain Adaptation Network (CDAN)~\cite{Long2018}, Pixel-level Domain Adaptation (PixelDA)~\cite{Bousmalis2017}, Unsupervised Image-to-Image translation (UNIT)~\cite{Liu2017}, Cycle-Consistent Adversarial Domain Adaption (CyCADA)~\cite{Hoffman2018}, and Generate to Adapt (GtA)~\cite{Sankaranarayanan2018}. We also used source- and target-only training as the lower and upper bound respectively, following the practice in~\cite{Bousmalis2017,Ganin2016}. 
 
\subsubsection{Quantitative Results}
\label{subsub:quanti-UDA}
The performance of domain adaptation for the different tasks is shown in Table~\ref{table:accuracy}. There are 4 scenarios and 7 tasks. Each scenario has bidirectional tasks for adaptation except LineMod. For LineMod, it is adapted from synthetic 3D image to real objects. For the same adaptation task, we cite the accuracy from the corresponding references, otherwise the accuracies for some tasks are obtained by training the open-source code provided by authors with suggested optimal parameters, for fair comparison. 

\begin{figure}
  \begin{subfigure}[t]{0.23\textwidth}
    \centering
    \includegraphics[width=0.9\textwidth]{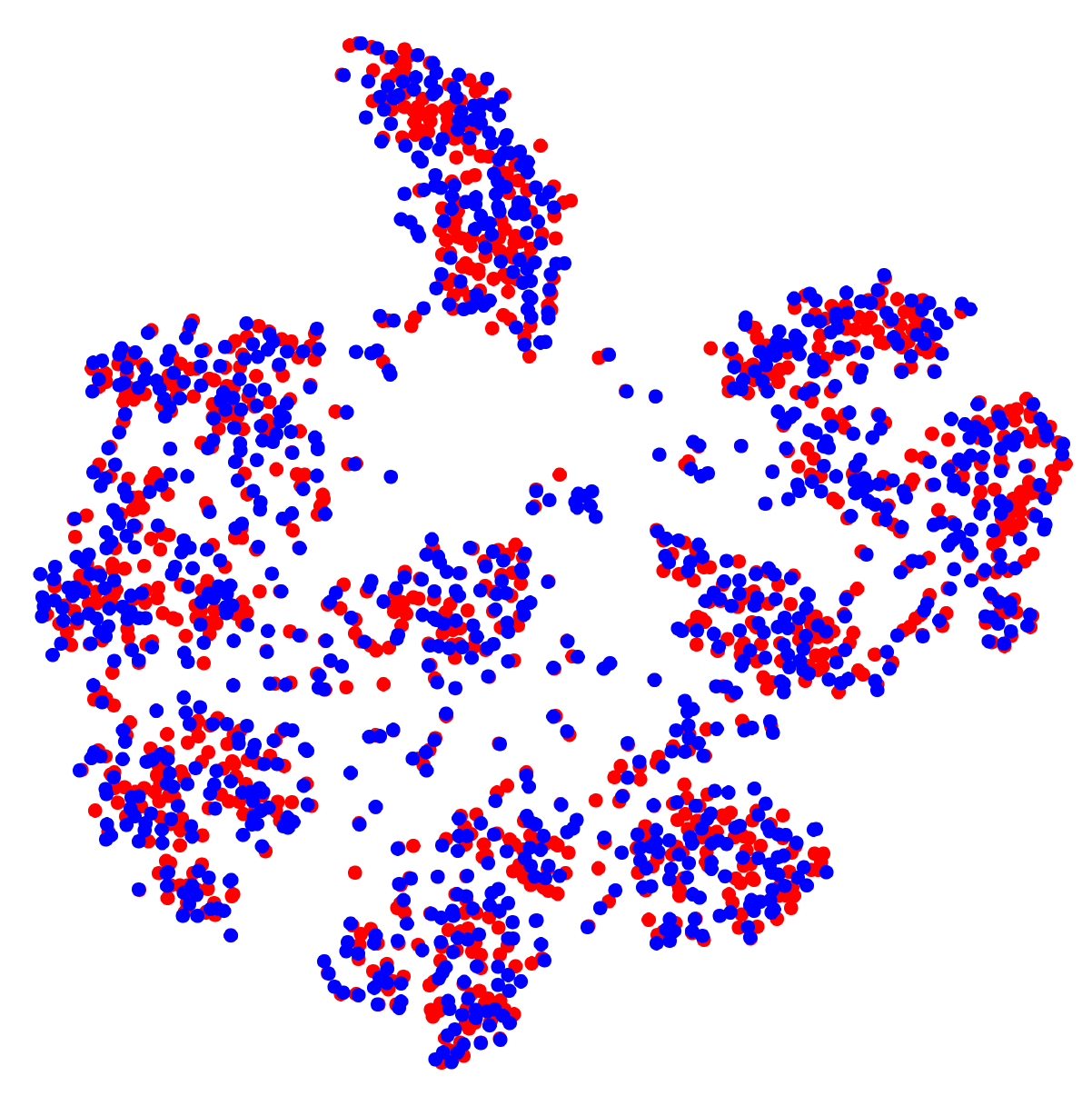}
    \caption{Task MNISTM$\rightarrow$MNIST}
    \label{fig:tsne_mnistm_mnist_z_h}
  \end{subfigure}
  \hfill
  \begin{subfigure}[t]{0.23\textwidth}
    \centering
    \includegraphics[width=0.9\textwidth]{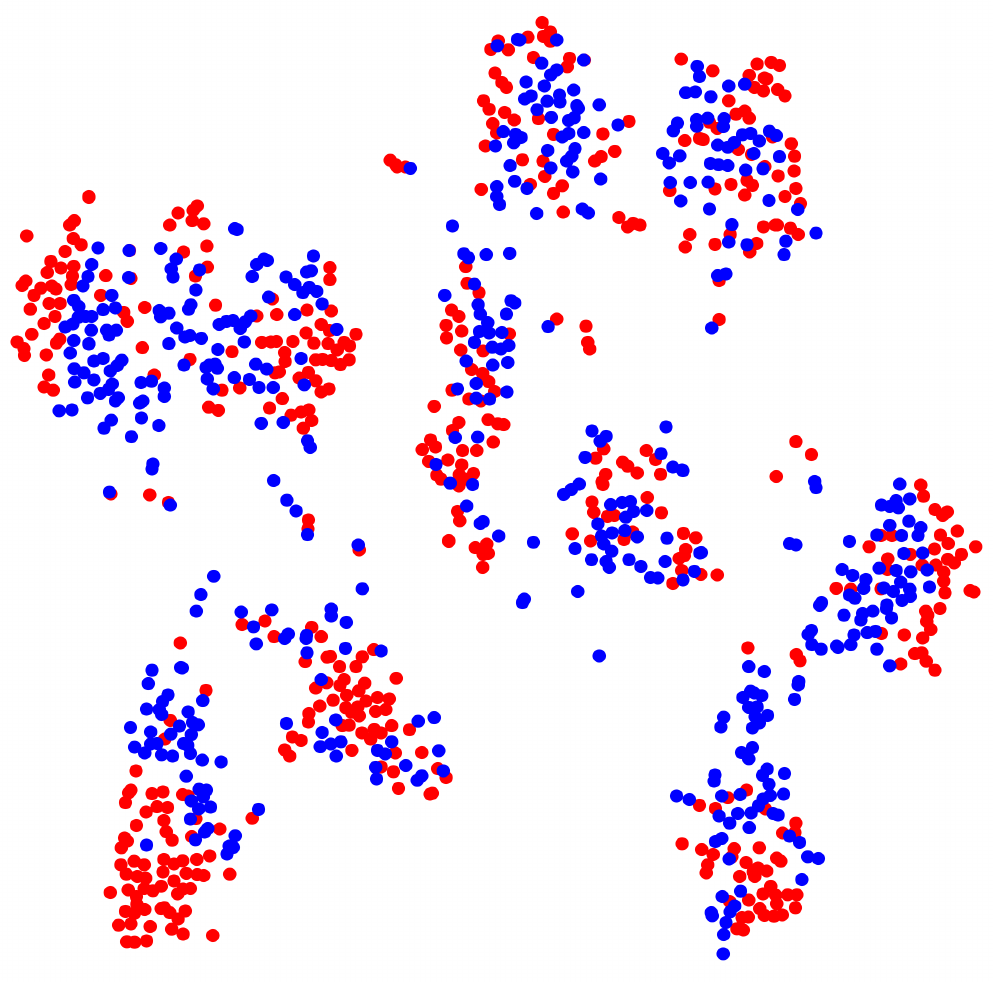}
    \caption{Task MNIST$\rightarrow$USPS.}
    \label{fig:tsne_mnist_usps_z_h}
  \end{subfigure}
  \caption{t-SNE visualization of cross-domain latent encodings $\bm{\ddot{z}}_{st}, \bm{\ddot{z}}_{ts}$. The $\bm{\ddot{z}}_{st}$ are in blue, and the $\bm{\ddot{z}}_{ts}$ in red.}
  \label{fig:tSNE}
\end{figure}

From Table~\ref{table:accuracy} we can see that the our method has a higher advantage compared with the baseline and the source-only accuracy, a little lower than the target-only accuracy from both adaptation directions. In comparison with other models, our model has a better performance for most of tasks. The CDLM has a higher adaptation accuracy for the scenarios with seemingly larger domain gap, such as MNIST$\rightarrow$MNISTM and Fashion$\rightarrow$FashionM. For the 3D scenario, the performance of our model is a little lower than PixelDA~\cite{Bousmalis2017}, but outperforms all the other compared methods. In PixelDA, the input is not only source image but also depth image pairs. It might be helpful for the generation. Besides, we visualize the t-SNE~\cite{Maaten2008} for latent encodings ($\bm{\ddot{z}}_{st}, \bm{\ddot{z}}_{ts}$) w.r.t the source and the target, respectively. Fig.~\ref{fig:tSNE} is the visualization for task MNISTM$\rightarrow$MNIST and MNIST$\rightarrow$USPS, and it shows that both are aligned well.

\subsubsection {Qualitative Results}
\label{subsub:visual-UDA}

\begin{figure}[]
    \centering
    \begin{subfigure}[t]{.48\textwidth}
        \centering
        \includegraphics[width=0.8\textwidth]{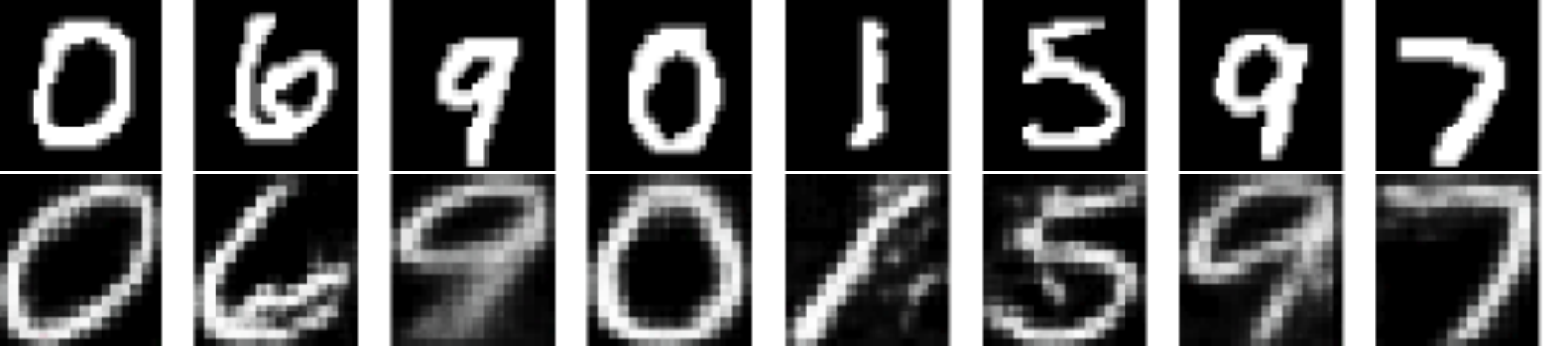} %
        \caption{USPS$\rightarrow$MNIST}
        \label{fig:usps_mnsit}
    \end{subfigure}
    \begin{subfigure}[t]{.48\textwidth}
        \centering
        \includegraphics[width=0.8\textwidth]{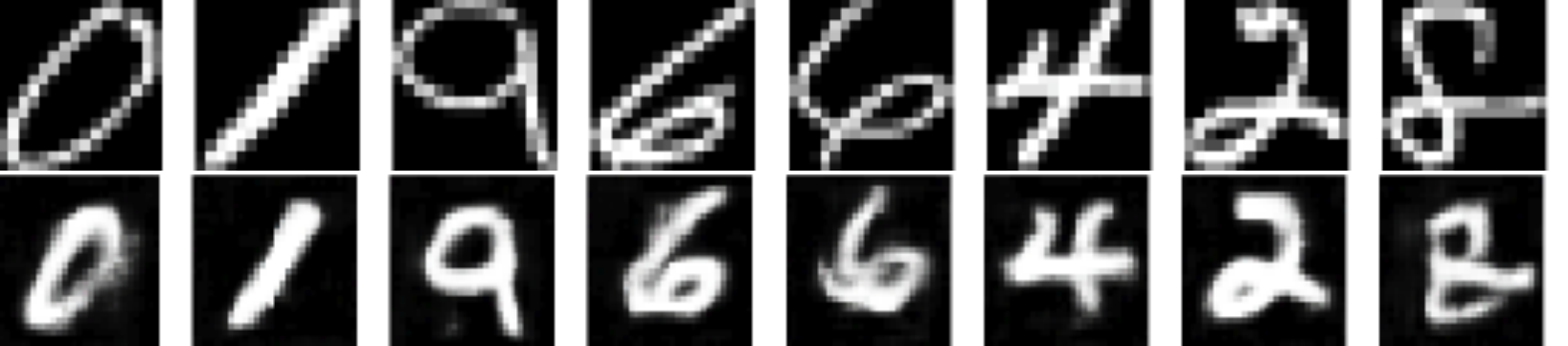} %
        \caption{MNIST$\rightarrow$USPS}
        \label{fig:mnist_usps}
    \end{subfigure}
    \begin{subfigure}[t]{.48\textwidth}
        \centering
        \includegraphics[width=0.8\textwidth]{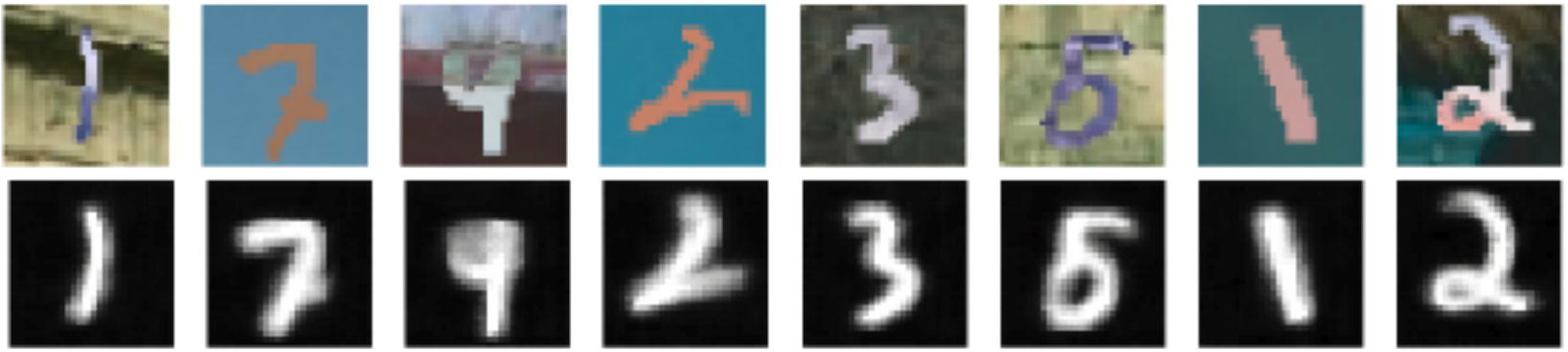}
        \caption{MNIST$\rightarrow$MNISTM}
        \label{fig:mnist_mnistm}
    \end{subfigure}
    \begin{subfigure}[t]{.48\textwidth}
        \centering
        \includegraphics[width=0.8\textwidth]{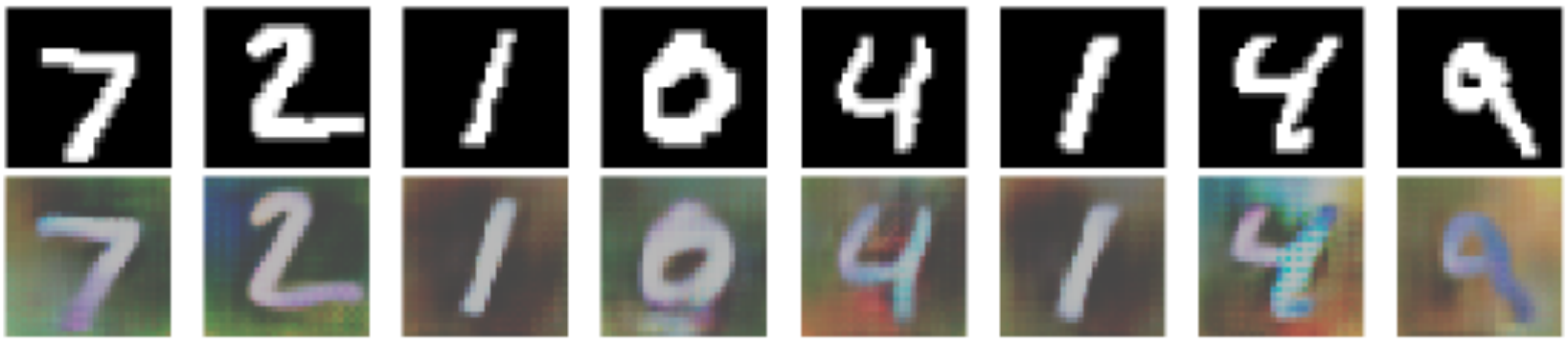}
        \caption{MNISTM$\rightarrow$MNIST}
        \label{fig:mnistm_mnist}
    \end{subfigure}
    \begin{subfigure}[t]{.48\textwidth}
        \centering
        \includegraphics[width=0.8\textwidth]{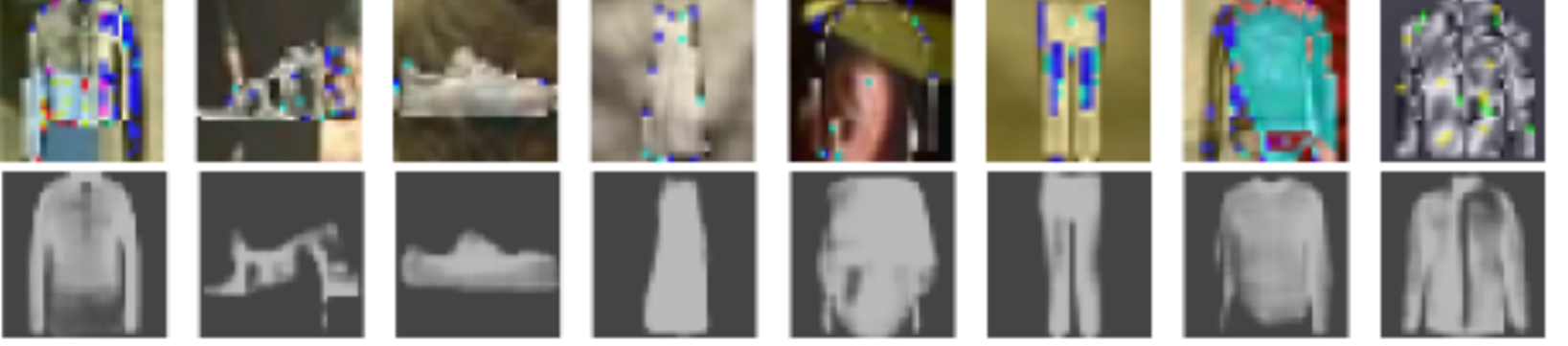}
        \caption{Fashion$\rightarrow$Fashionm}
        \label{fig:fashion_fashionm}
    \end{subfigure}
    \begin{subfigure}[t]{.48\textwidth}
        \centering
        \includegraphics[width=0.8\textwidth]{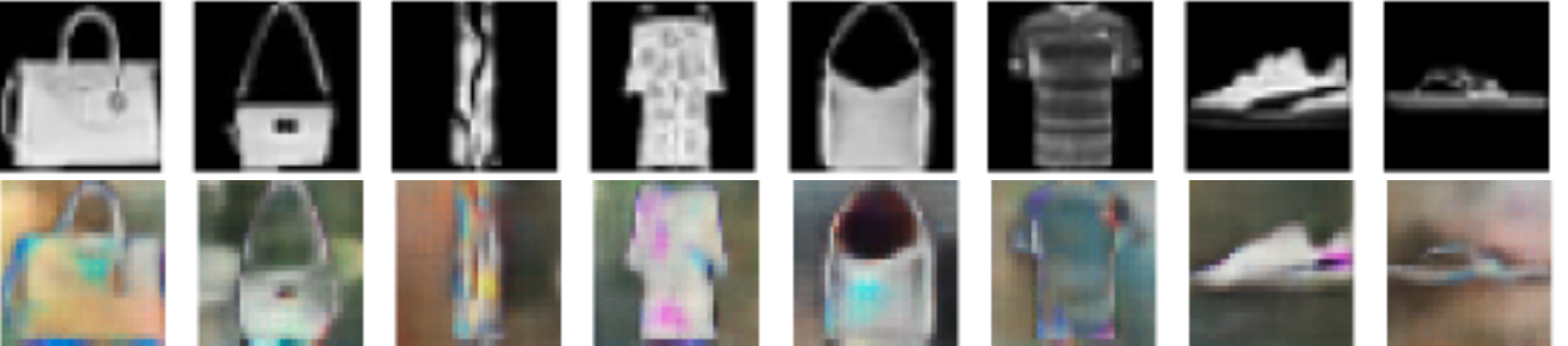}
        \caption{Fashionm$\rightarrow$Fashion}
        \label{fig:fashionm_fashion}
    \end{subfigure}
    \caption{Visualization for the adaptations. 6 different tasks are illustrated. For each task, the first row shows target images and the second row shows the adapted images with source-like style.}
    \label{fig:digitsandfashion}   
\end{figure}

Our model can give the visualization of the adaptation. Fig.~\ref{fig:digitsandfashion} is the visualization for the digits and Fashion adaptation respectively. For the scenario of MNIST and USPS, the generation for the task USPS$\rightarrow$MNIST is shown in Fig.~\ref{fig:usps_mnsit}. The target MNIST is transferred to the source USPS style well, meanwhile it keeps the correspondent content (label). For example, the digit `1' in MNIST become more leaned and `9' more flatten. Also in Fig.~\ref{fig:mnist_usps}, the target USPS becomes the MNIST style. For the scenario of MNIST$\rightarrow$MNISTM, our proposed model can remove and add the noise background well for adaptation.

\begin{figure}  
  \centering
  \includegraphics[width=0.45\textwidth]{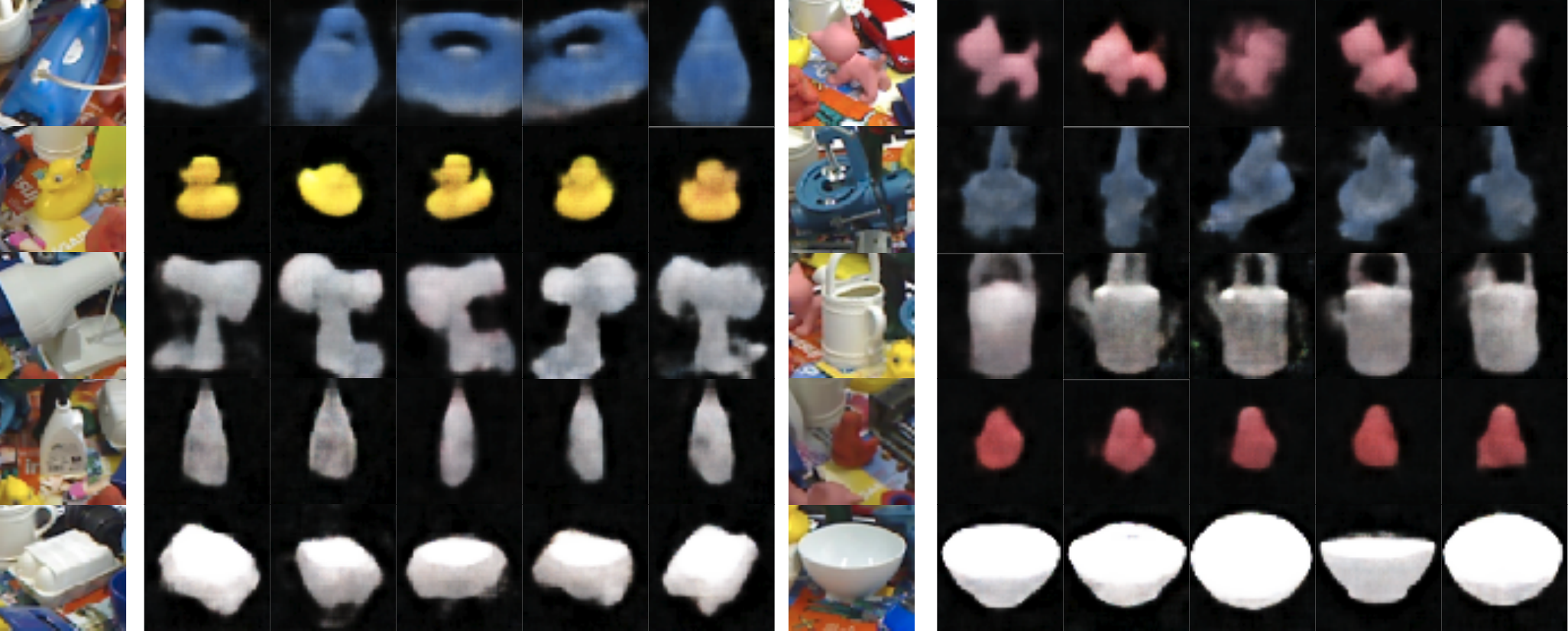}
  \caption{Linemod 3D Synthetic$\rightarrow$Real. For a query image (on the left), different adaptation images (to the right) with various poses can be generated.}
  \label{fig:linemod}
\end{figure}

For the scenario of Fashion, the fashion items have more complicated texture and content. In addition, the noisy backgrounds pollute the items randomly, for example, different parts of a cloth are filled with various colors. For visualizations, specifically, Fig.~\ref{fig:fashion_fashionm} is for the task Fashion $\rightarrow$ FashionM. The proposed model can remove the noisy background and maintain the content. On the other hand, Fig.~\ref{fig:fashionm_fashion} shows that the original Fashion images are added with similar noisy background as the source. This is promising for a better adaptation performance. 

For  Linemod3D, the real objects images with different backgrounds are transferred to the synthetic 3D images with black background. Due to the 3D style, the generation of the target gives different poses. For example in Fig.\ref{fig:linemod}, different poses of the iron object are obtained for different trials.


\begin{figure*}[!htbp]
    \centering
    \begin{subfigure}{.49\textwidth}
        \centering
        \includegraphics[width=1.0\textwidth]{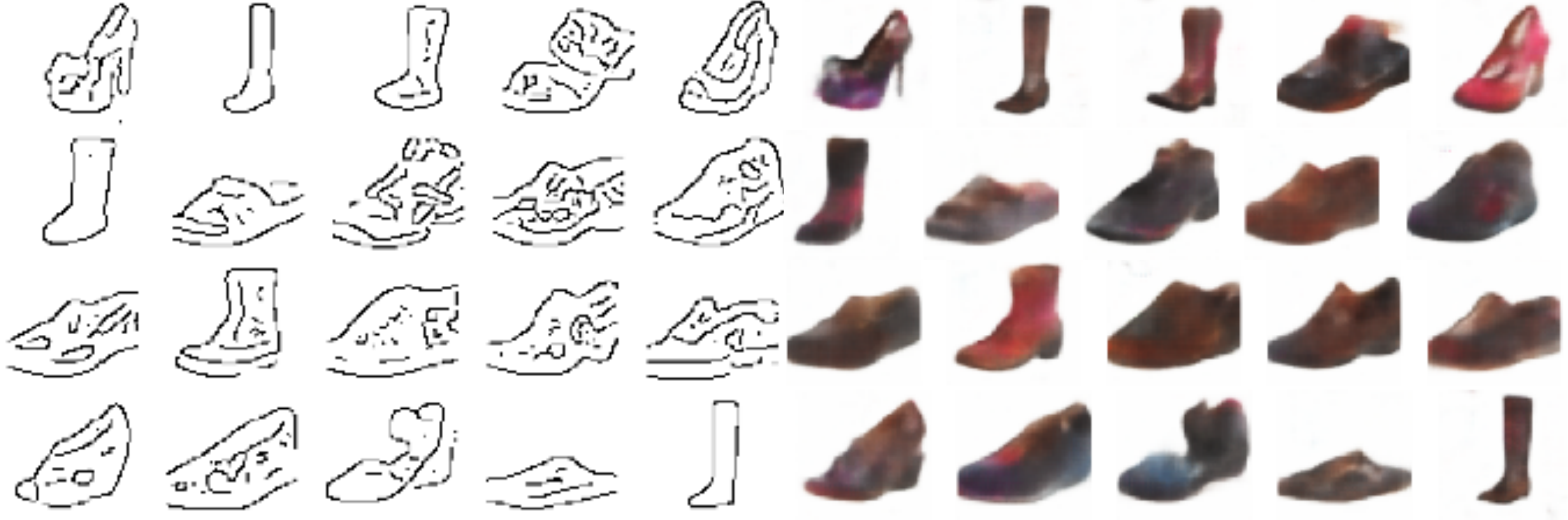} %
        \caption{``edge'' to ``shoes''}
        \label{fig:edge_shoes}
    \end{subfigure} \hfill
    \begin{subfigure}{.49\textwidth}
        \centering
        \includegraphics[width=1.0\textwidth]{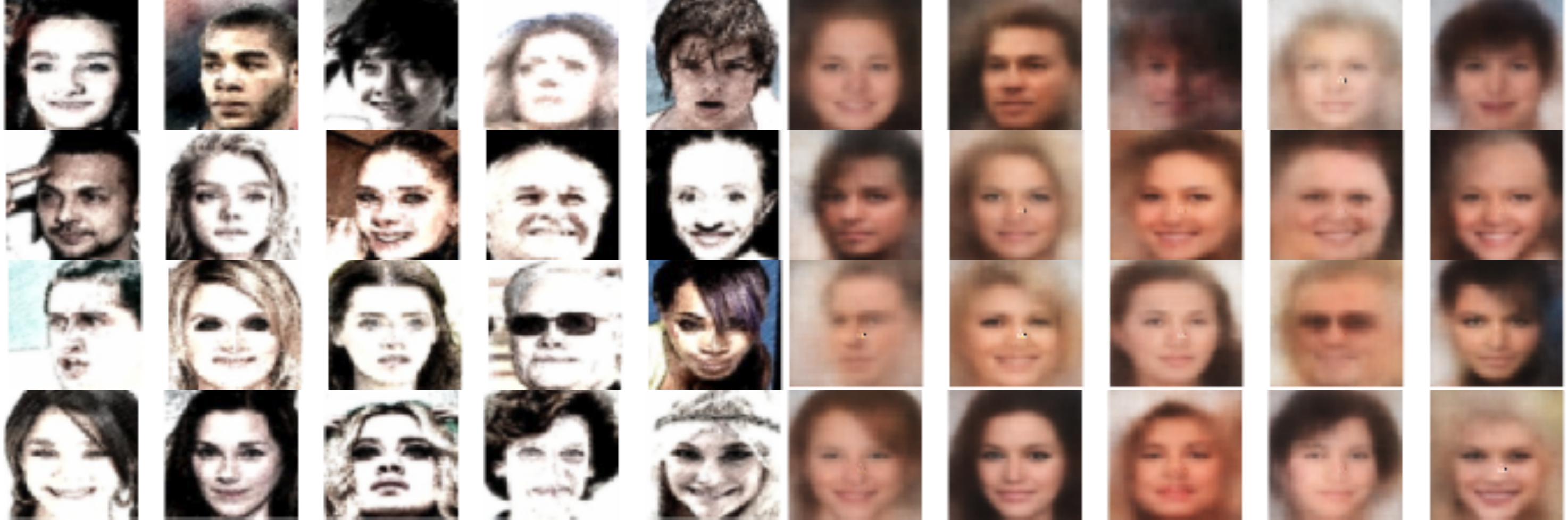}  %
        \caption{``sketch'' to ``face''}
        \label{fig:sketch_face}
    \end{subfigure}
    \begin{subfigure}{.49\textwidth}
        \centering
        \includegraphics[width=1.0\textwidth]{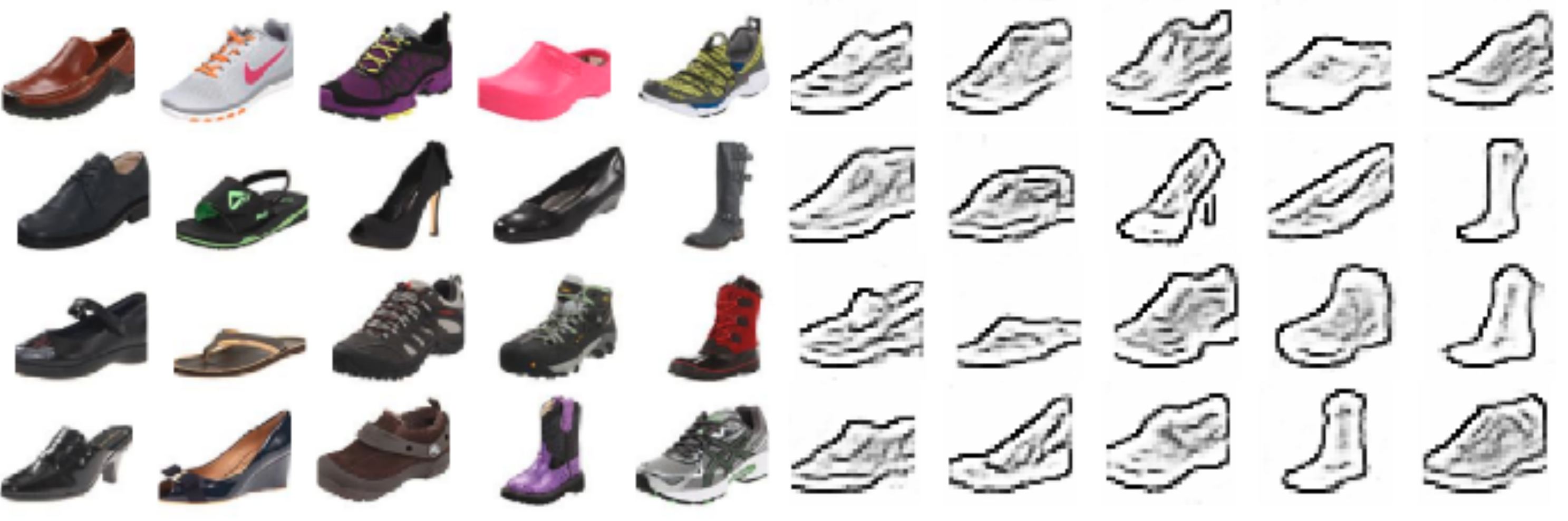} %
        \caption{``shoes'' to ``edge''}
        \label{fig:shoes_edge}
    \end{subfigure} \hfill
    \begin{subfigure}{.49\textwidth}
        \centering
        \includegraphics[width=1.0\textwidth]{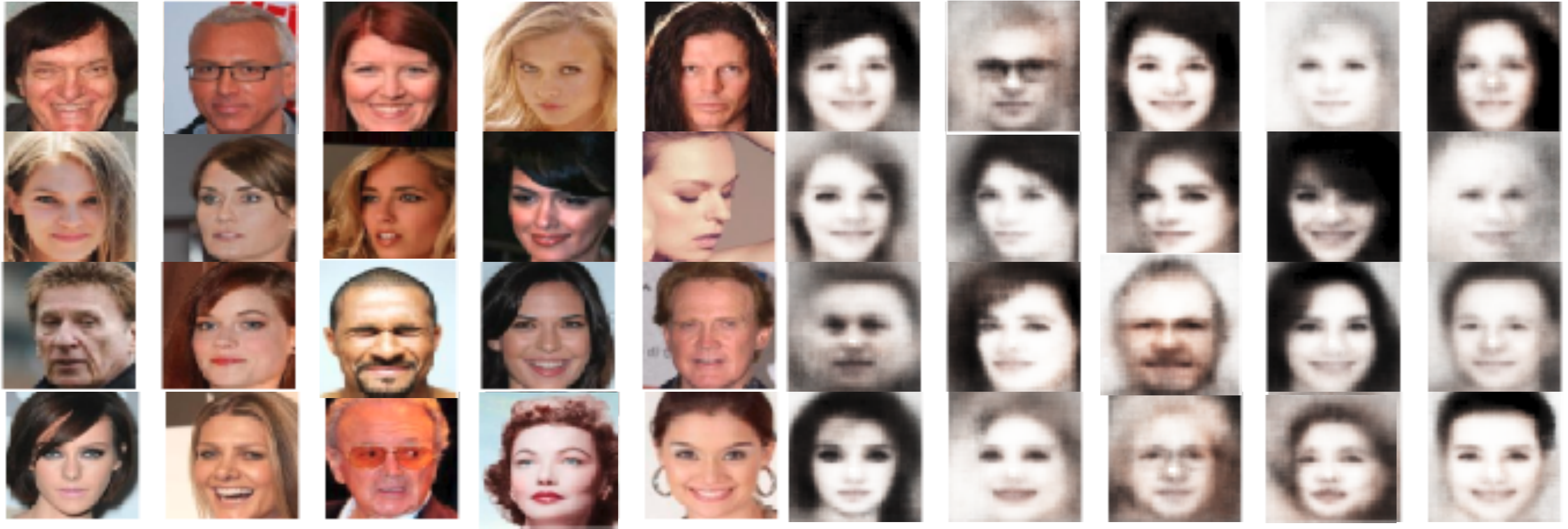} %
        \caption{``face'' to ``sketch''}
        \label{fig:face_sketch}
    \end{subfigure}
    \caption{Visualization of cross-domain image mapping. 
    }
    \label{fig:imagetrans}   
\end{figure*}

\subsection{Cross-Domain Image Mapping}
\label{UIT}

The proposed model also can be used for the cross-domain image mapping. 

Fig.~\ref{fig:imagetrans} gives a demonstration of the image style translation. Specifically, 
for ``shoes'' and ``edges'' in Fig.~\ref{fig:edge_shoes} and~\ref{fig:shoes_edge}, we can see that the proposed model can translate ``edges'' to its counterpart quite well. The translation is stochastic -- an ``edge'' pattern can be used to generate ``shoes'' in  different colors with different trials. For the more challenging ``face'' and ``sketch'' translations, the proposed model also performs well. The generations have some variations compared with the original images. In general,  our method can generate realistic translated images. However, we find that compared with the translation from sketches to real images, the reverse task seems harder. For example, when a face image is given, the generated sketch loses some details. The reason may be the low-level feature is neglected when the deep feature acts as the condition.  

For further evaluation, quantitative performance is evaluated for image mapping.  SSIM~\cite{Wang2004}, MSE, and PSNR are used for the evaluation. The results are shown in Table~\ref{table:image_mapping_quantitative_results}. We can see that our model outperforms E-CDRD~\cite{Liu2018}, which learns a disentangled latent encoding for the source and the target for domain adaptation. Meanwhile, it matches the performance of StarGAN~\cite{Choi2018}, which is designed for multi-domain image translation. The result shows that our model can map cross-domain images well compared to these prior works.

\begin{table}[!thbp]
  \centering
  \caption{\small Performance for image mapping.}
  \resizebox{0.33\textwidth}{!}{
  \begin{tabular}{|c|c|c|c|}
    \hline
    \multirow{2}{*}{Models} & \multicolumn{3}{|c|}{``Sketch'' to ``Face''} \\
    \cline{2-4}
    & SSIM  &  MSE & PSNR  \\
    \hline \hline
    E-CDRD~\cite{Liu2018} & 0.6229  & 0.0207 & 16.86 \\
    StarGAN~\cite{Choi2018} & \textbf{0.8026} & 0.0142 &  19.04 \\
    \hline
    CDLM & 0.7961 & \textbf{0.0140} &  \textbf{19.89}  \\
    \hline
  \end{tabular}}
  \label{table:image_mapping_quantitative_results}
\end{table}

In addition, we also conduct the classification to evaluate the translation performance. We take shoes as an example which are labeled to 4 different classes. The recognition accuracy of our proposed model for task shoes$\rightarrow$edge is 0.953, which is higher than the results of PixelDA (0.921) and UNIT (0.916) respectively.

\subsection {Model Analysis}
\label{subsec:ablation}

\noindent
\textbf {The effect of encoder settings -- depth and different $\gamma_1, \gamma_2$:} In our model, the deep features are utilized to cross-modulate the transfer latent encoding. Therefore the deep feature is an important factor in our framework and is influenced by the depth of the encoder. During the experiments, we use MNIST $\rightarrow$ USPS and Fashion $\rightarrow$ FashionM as the evaluation tasks. For the first one, they have different content, but with the same background. The second task is a totally different scenario, the images have the same content but different background. The outputs of different encoder layers ($k > 3$) are used for the experiments. 

\begin{table} [!hbtp]
    \centering
    \caption{\small Adaptation accuracy with different layer depth for Tasks MNIST $\rightarrow$ USPS and Fashion $\rightarrow$ FashionM. }
    \resizebox{0.4\textwidth}{!}{
        \begin{tabular}{|c|c|c|c|}
            \hline
            \small Tasks/Layers & \small Conv4  & \small Conv5  &  \small Conv\textsubscript{last}  \\
            \hline \hline
            \small MNIST $\rightarrow$ USPS & 0.954  & 0.956  & 0.961   \\
            \small Fashion $\rightarrow$ FashionM& 0.890 & 0.905  &  0.913  \\
            \hline
    \end{tabular}}
    \label{table:layers} 
\end{table}

\begin{table}[!htbp]
    \centering
    \caption{\small Adaptation accuracy with different ($\gamma_1, \gamma_2$) for Tasks MNIST $\rightarrow$ USPS and Fashion $\rightarrow$ FashionM.}
    \resizebox{0.48\textwidth}{!}{
    \begin{tabular}{|c|c|c|c|c|c|}
            \hline
            \small Tasks / ($\gamma_1,\gamma_2$) & (\small 0.1,1.0)  &  (\small 0.5,0.5)  &  (\small 0.9,0.1) & (\small 1.0,0.1) & (\small 1.0, 0)\\
            \hline \hline
            \small MNIST $\rightarrow$ USPS & {0.320}  & {0.723}  & 0.961 & 0.961 & 0.961\\
            \small Fashion $\rightarrow$ FashionM & {0.226} & {0.513}  &  0.912 & 0.913 & 0.913 \\
            \hline
    \end{tabular}}
    \label{table:gamma}
\end{table}

\begin{table*}[!htbp]
    \centering
    \caption{\small Evaluation on the effect of unsupervised consistency metrics. The recognition accuracy is shown for four tasks in the unsupervised domain adaptation scenario. Our model is on the last row with both the $\mathcal{L}_c^s$ and $\mathcal{L}_c^t$, which achieves the best performance.}
    \resizebox{0.85\textwidth}{!}{
        \begin{tabular}{|l|c|c|c|c|}
            \hline
            Model/Tasks & MNIST$\rightarrow$USPS & USPS$\rightarrow$MNIST & Fashion$\rightarrow$FashionM & FashionM$\rightarrow$Fashion \\
            \hline \hline
            CDLM w/o $\mathcal{L}_c$ & 0.635 & 0.683 & 0.646 & 0.672  \\
            CDLM+$\mathcal{L}_c^t$ & 0.689 & 0.695 & 0.682 & 0.691  \\
            CDLM+$\mathcal{L}_c^s$ & 0.951 & 0.980 & 0.912 & 0.915  \\
            CDLM+$\mathcal{L}_c^s$+$\mathcal{L}_c^t$  & 0.961 & 0.983 & 0.913 & 0.922  \\
            \hline   
    \end{tabular}}
    \label{table:wo_sim}
\end{table*}

As the result (Table~\ref{table:layers}) shows, a higher accuracy is achieved when more layers are used to extract the deep representations. The accuracy gain of the task MNIST $\rightarrow$ USPS is lower than that of Fashion $\rightarrow$ FashionM. This is expected as features extracted by higher layers would normally eliminate lower-level variations between domains, such as change of background and illumination in the images. 

For $\gamma_1, \gamma_2$, we fixed the last convolutional layer for the deep representations and evaluate different values. From Table~\ref{table:gamma}, we can see that the performance drops down significantly with a smaller $\gamma_1$ compared with $\gamma_2$, and increased with a larger $\gamma_1$. The performance seems to be stabilized when  $\gamma_1$ is greater than 0.9 while $\gamma_2$ remains 0.1. Following the standard VAE, we keep the noise $\bm \epsilon$ ($\gamma_2\neq 0$) in the evaluations. Meanwhile, our model works well even when $\gamma_2=0$. These results suggest the deep representation plays a crucial role in the cross-domain modulation.

\begin{figure}[!htbp]
    \centering
    \includegraphics[width=0.75\columnwidth]{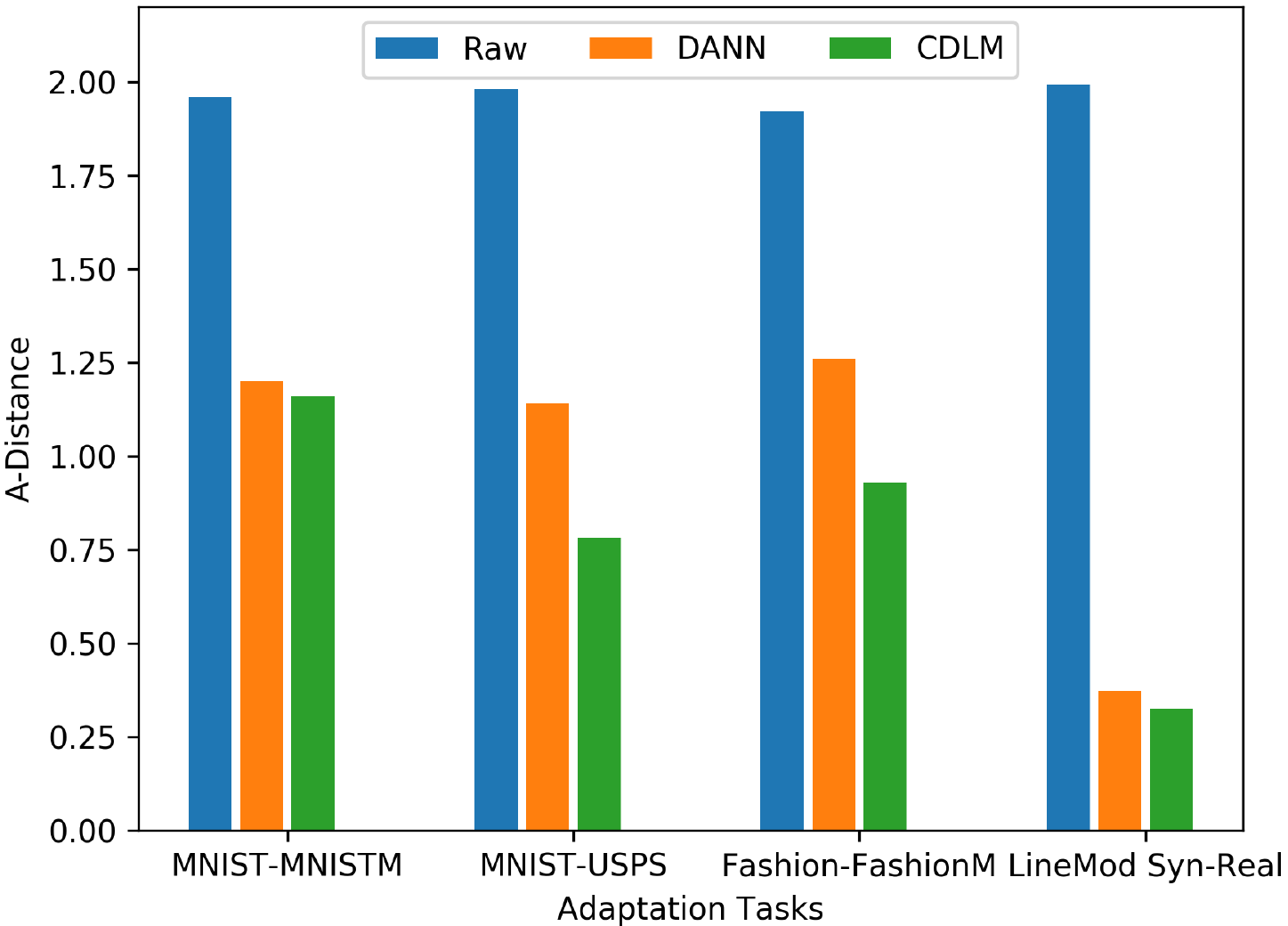}
    \caption{$\mathcal{A}$-distances comparison for four tasks.}
    \label{fig:A-distance}
\end{figure}

\noindent
\textbf{$\mathcal{A}$-Distance:} In a theoretical analysis of the domain discrepancy~\cite{Ben-David2007}, Ben-David \textit{et al.} suggests that $\mathcal{A}$-distance can be used as a measure for the domain discrepancy. As the exact $\mathcal{A}$-distance is intractable, a proxy is defined as $\hat{d}_{\mathcal{A}} = 2(1 - 2\epsilon)$, where $\epsilon$ is the generalization error of a binary classifier (e.g. kernel SVM) trained to distinguish the input's domain (source or target). Following the protocol of~\cite{Long2015,Peng2020}, we calculate the $\mathcal{A}$-distance on four adaptation tasks under the scenarios of Raw features, DANN features, and CDLM features respectively. The results are show in Fig.~\ref{fig:A-distance}. We observe that both the DANN and CDLM reduce the domain discrepancy compared with the Raw images scenario, and the $\mathcal{A}$-distance of CDLM is smaller than the DANN's. This demonstrates that it is harder to distinguish the source and the target by the CDLM generations.

\begin{figure}[!htbp]
    \centering
    \includegraphics[width=0.75\columnwidth]{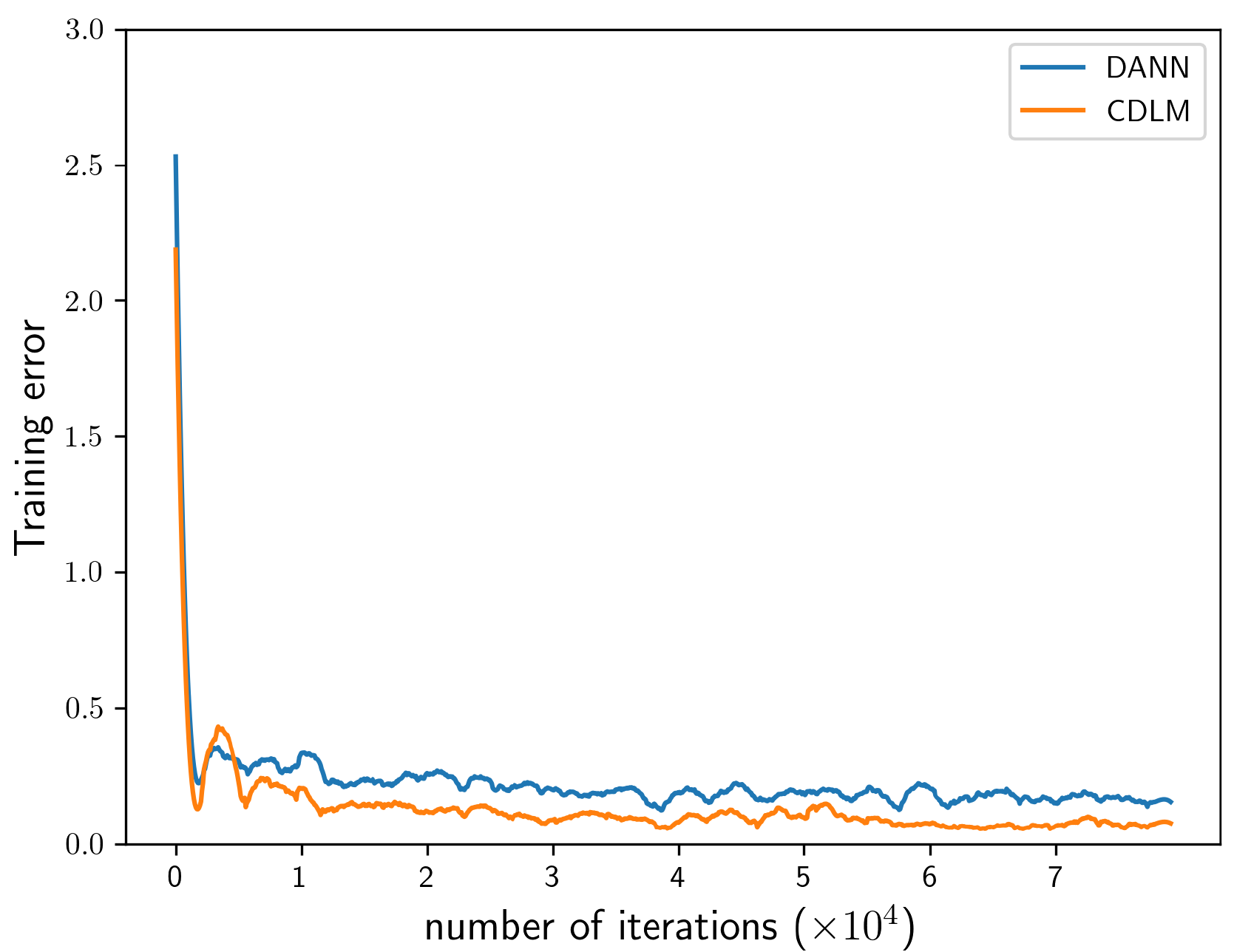}
    \caption{Convergence of CDLM compared with DANN.}
    \label{fig:convergence}
\end{figure}

\noindent
\textbf{Convergence:} We also conduct the convergence experiment with training error on task MNSIT-USPS to evaluate our model. As shown in the Fig.~\ref{fig:convergence}, our model has a better convergence than DANN, thought there are some oscillations at the beginning of the training. In addition, the error of CDLM is lower that the DANN, which demonstrate that CDLM has a better adaptation performance. This is consistent with the adaptation performance in Table~\ref{table:accuracy}.     

\noindent
\textbf {The effect of unsupervised consistency metrics:} In our model, two unsupervised consistency metrics are added for generation in good effects. The adaptation accuracy is used for evaluation. Table~\ref{table:wo_sim} is the results for the four different tasks. The performance w/o $\mathcal{L}_c$ is dropped down because the decoder cannot generate realistic cross-domain images. $\mathcal{L}_c^t$ connects outputs generated from the $\bm h_t$ and $\ddot{\bm z}_{ts}$ only for the target, which improves the performance slightly. Meanwhile, we can see that the $\mathcal{L}_c^s$ loss boosts the accuracy for adaptation significantly, which connects the two domains with the generations by the $\bm h$. Finally, the scenario with both $\mathcal{L}_c^s$ and $\mathcal{L}_c^t$ gives the best performance in all four tasks. It bridges both the $\bm h$ and $\ddot{\bm z}$ between the two domains.

\section{Conclusion} \label{sec:conclusion}

In this paper, we have presented a novel variational cross-domain transfer learning model with cross modulation of deep representations from different domains. A shared transfer latent space is introduced, and the reparameterization transformation is modified to enforce the connection between domains. Evaluations carried out in unsupervised domain adaptation and image translation tasks demonstrate our model's competitive performance. Its effectiveness is also clearly shown in visual assessment of the adapted images, as well as in the alignment of the latent information as revealed by visualization using t-SNE. Overall, competitive performance has been achieved by our model  despite its relative simplicity. 

For future work, we intend to further improve our variational transfer learning framework and use it for heterogeneous, multi-domain transfer tasks.  


{\small
\bibliographystyle{ieee_fullname}
\bibliography{cdlm}
}


\end{document}